\documentclass[10pt,twocolumn,letterpaper]{article}

\usepackage{cvpr}
\usepackage{times}
\usepackage{epsfig}
\usepackage{graphicx}
\usepackage{amsmath}
\usepackage{amssymb}
\usepackage{mathrsfs}
\usepackage{subfigure}

\long\def\symbolfootnote[#1]#2{\begingroup
\def\thefootnote{\fnsymbol{footnote}}\footnote[#1]{#2}\endgroup}

\usepackage[pagebackref=true,breaklinks=true,letterpaper=true,colorlinks,bookmarks=false]{hyperref}

 \cvprfinalcopy 

\def\cvprPaperID{1679} 

\ifcvprfinal\fi
\begin{document}

\newcommand{\authorfont}{\fontsize{11pt}{\baselineskip}\selectfont}

\title{Deep Structured Scene Parsing by Learning with Image Descriptions}

\author{ \authorfont{Liang Lin$^1$, Guangrun Wang$^{1}$, Rui Zhang$^{1}$, Ruimao Zhang$^{1}$, Xiaodan Liang$^1$, Wangmeng Zuo$^2$}\\
{\small $^1$School of Data and Computer Science, Sun Yat-sen University, Guangzhou, China} \\
{\small $^2$School of Computer Science and Technology, Harbin Institute of Technology, China}\\
{\tt\small linliang@ieee.org; r.m.zhang1989@gmail.com; cswmzuo@gmail.com.}
}

\maketitle

\begin{abstract}
This paper addresses a fundamental problem of scene understanding: How to parse the scene image into a structured configuration (i.e., a semantic object hierarchy with object interaction relations) that finely accords with human perception. We propose a deep architecture consisting of two networks: i) a convolutional neural network (CNN) extracting the image representation for pixelwise object labeling and ii) a recursive neural network (RNN) discovering the hierarchical object structure and the inter-object relations. Rather than relying on elaborative user annotations (e.g., manually labeling semantic maps and relations), we train our deep model in a weakly-supervised manner by leveraging the descriptive sentences of the training images. Specifically, we decompose each sentence into a semantic tree consisting of nouns and verb phrases, and facilitate these trees discovering the configurations of the training images. Once these scene configurations are determined, then the parameters of both the CNN and RNN are updated accordingly by back propagation. The entire model training is accomplished through an Expectation-Maximization method. Extensive experiments suggest that our model is capable of producing meaningful and structured scene configurations and achieving more favorable scene labeling performance on PASCAL VOC 2012 over other state-of-the-art weakly-supervised methods.

\end{abstract}

\section{Introduction}

Scene understanding started with the goal of creating systems that can infer meaningful configurations (e.g., parts, objects and their compositions with relations) from imagery like humans~\cite{DBLP:ImageParsing-Attribute}. In computer vision research, significant progresses have been made in semantic scene labeling / segmentation (i.e., assigning the label for each pixel of the scene image)~\cite{DBLP:SemanSeg-1}\cite{DBLP:SemanSeg-3}\cite{DBLP:FCnetwork}\cite{DBLP:RecursiveContext}. However, the problem of structured scene parsing (i.e., producing meaningful scene configurations) remains a challenge due to the following difficulties.

\begin{itemize}

\item The representations of nested hierarchical structure in scene images are often ambiguous, \eg, a configuration may have more than one way of parsing. Conducting these parsing results to finely accord with human perception is an interesting yet fundamental problem.

\item Training a scene parsing model usually relies on very expensive manual annotations, \eg, including semantic maps and structured configurations.

\end{itemize}

\begin{figure}[t]
\centering
\includegraphics[width=3.3in]{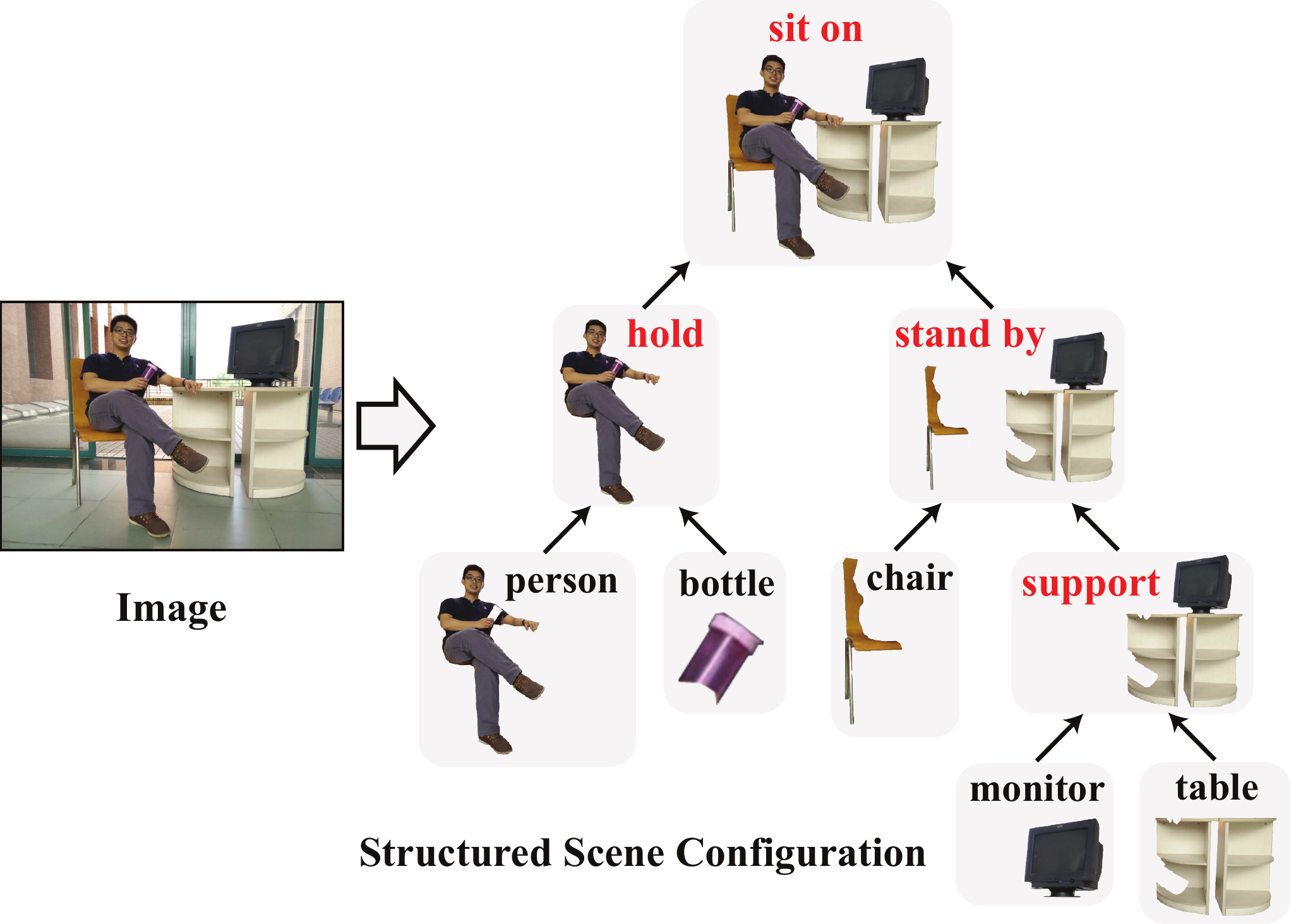}
\caption{An illustration of our structured scene parsing. An input scene image is automatically parsed into a hierarchical configuration that comprises hierarchical semantic objects (black labels) and the interaction relations (red labels) of objects.}
\label{fig:application}
\vspace{-4mm}
\end{figure}

To address these above issues, we develop a novel deep neural network architecture that automatically parses an input scene into a structured and meaningful configuration. Fig.~\ref{fig:application} shows an illustration of our structured scene parsing, where our model identifies salient semantic objects in the scene and generates the hierarchical scene structure with the interaction relations among objects. Our model is inspired by the effectiveness of two widely successful deep learning techniques: convolutional neural networks (CNNs)~\cite{DBLP:AlexNet}\cite{DBLP:FCnetwork} and recursive neural networks (RNNs)~\cite{DBLP:Recursive_Socher}. The former category of models is widely applied for generating powerful feature representations in various vision tasks such as image classification and object recognition. Meanwhile, the RNN models (such as \cite{DBLP:Recursive_Socher}\cite{DBLP:RecursiveContext}\cite{DBLP:RecursiveContext2}) have demonstrated as an effective class of models for predicting hierarchical and compositional structures in image and natural language understanding~\cite{DBLP:RNN-NLP}. One important property of RNNs is the ability to recursively learn the representations in a semantically and structurally coherent way. In our deep CNN-RNN architecture, the CNN and RNN models are collaboratively integrated for accomplishing the scene parsing from complementary aspects. We utilize the CNN to layerwise extract features from the input scene image and generate the representations of semantic objects. Then, the RNN is sequentially stacked based on the CNN feature representations, generating the structured configuration of the scene.

On the other hand, to avoid relying on the elaborative annotations, we propose to train our CNN-RNN model by leveraging the image descriptions. Our approach is partially motivated but different with the recently proposed methods for image-sentence embedding~\cite{DBLP:VisualAlign-feifei}\cite{DBLP:TellMeShowYou}. In particular, we distill knowledge from the sentence descriptions for discovering scene structural configurations.

In the initial stage, we decompose each sentence into a normalized semantic tree consisting of nouns and verb phrases by using a standard parser~\cite{DBLP:conf/acl/SocherBMN13} and the WordNet\cite{wordnet}. Afterward, based on these semantic trees and their associated scene images, we train our model by developing an Expectation-Maximization method. Specifically, the semantic tree facilitates discovering the latent scene configuration in the two following aspects. i) The entities (\ie, nouns) determine the object category labels existing in the scene, and ii) the relations (\ie, verb phrases) over the entities assist to produce the scene hierarchy and object interactions. The two proportions of knowledge are incorporated into our learning objective together with the CNN and the RNN, respectively. Therefore, once the scene configuration is fixed, the parameters of the two neural networks are updated accordingly by the back propagation.

The main contributions of our work are summarized as follows. i) We present a novel CNN-RNN framework for generating meaningful and hierarchical scene representations, which gains a deeper understanding of the objects in the scene compared to traditional scene labeling. The integration of CNN and RNN models is general to be extended to other high-level computer vision tasks. ii) We present a EM-type training method by leveraging text descriptions that associate with the training images. This method is cost-effective yet beneficial to introducing rich contexts and semantics. iii) Our extensive experiments on PASCAL VOC 2012 demonstrate that the parsed scene representations are useful for scene understanding and our generated semantic segmentations are more favorable than those by other weakly-supervised scene labeling methods.


\section{Related Work}

Scene understanding is arguably considered as the most fundamental problem in computer vision, which actually involves several tasks of different level. In current research, a myriad of different methods focus on what general scene type the image shows (classification)~\cite{DBLP:VisualAttr}\cite{DBLP:Multi-Class}\cite{DBLP:fine-grained}, what objects and their locations are in a scene (semantic labeling or segmentation)~\cite{DBLP:Seg-03}\cite{DBLP:Seg-09}\cite{DBLP:Seg-15}\cite{tighe2014scene}. These methods, however, ignore or over-simplified the compositional object representations and would fail to gain a deeper scene understanding.


Meanwhile, as a higher-level task, structured scene parsing has also attracted much attention. A pioneer work was proposed by Tu et al.,~\cite{tu2005image}, in which they mainly focused on faces and texture patterns by a Bayesian inference framework. In \cite{DBLP:ImageParsing-Attribute}, Han et al., proposed to hierarchically parse the indoor scene images by developing a generative grammar model. A hierarchical model was proposed in \cite{zhu2012recursive} to represent the image recursively by contextualized templates at multiple scales, and the rapid inference was realized based on dynamic programming. Ahuja et al.,~\cite{ahuja2008connected} developed a connected segmentation tree for object and scene parsing. Some other related works~\cite{silberman2012indoor}\cite{gupta2013perceptual} investigated the approaches for RGB-D scene understanding, achieving impressive results.


With the resurgence of neural network models, the performances of scene understanding have been improved substantially. The representative works, the fully convolutional network (FCN)~\cite{DBLP:FCnetwork} and its extensions~\cite{DBLP:CNN-CRF}, demonstrate effectiveness in pixel-wise scene labeling. A recurrent neural network model was proposed in \cite{DBLP:CRF-RNN}, which improves the segmentation performance by incorporating the mean-field approximate inference, and similar idea was also explored in \cite{DBLP:CNN-MRF}. For the problem of structured scene parsing, recursive neural networks (RNNs) were studied in \cite{DBLP:Recursive_Socher}\cite{DBLP:RecursiveContext2}. For example, Socher et al.~\cite{DBLP:Recursive_Socher} proposed to predict hierarchical scene structures by using a max-margin RNN model. The differences between these existing RNN-based parsing models and our model are two-fold. First, they mainly focused on parsing only the semantic entities (\eg, buildings, bikes, trees) and the scene configurations generated by ours include not only the objects but also the interaction relations of objects. Second, we incorporate convolutional feature learning into our deep model for joint optimization.


\begin{figure*}[t!]
\centering
\includegraphics[width=6.8 in]{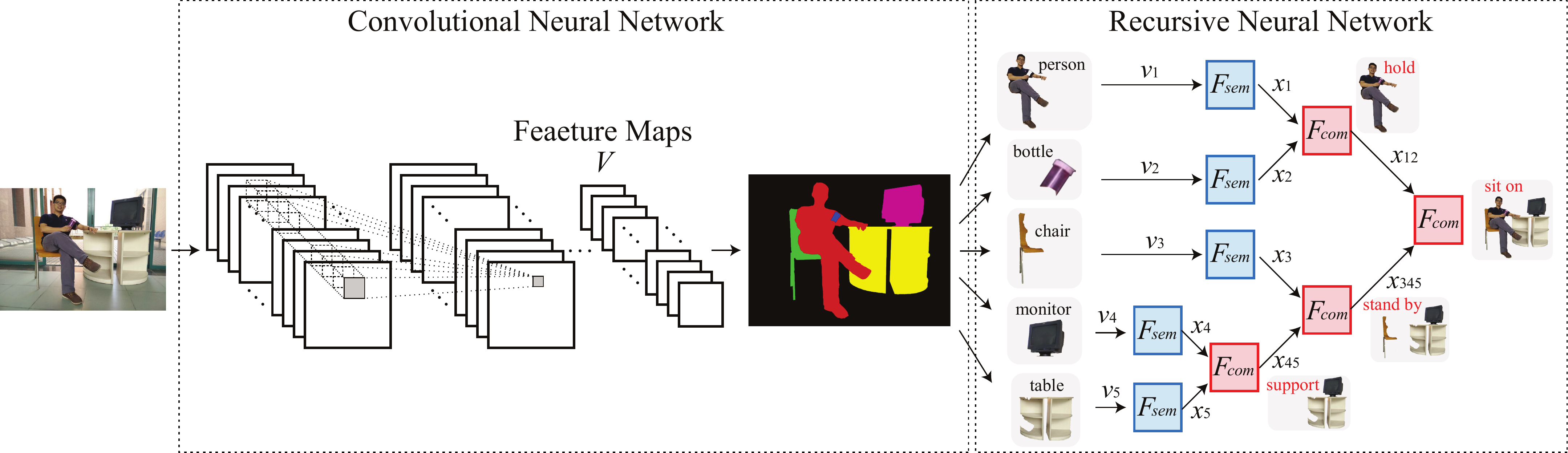}
\caption{A glance into our proposed CNN-RNN architecture for structured scene parsing. The CNN takes the image as input and produces the pixel-wise semantic score map. Then the pixels with the same label are grouped into a semantic object, and we can obtain the feature representations (i.e., $v_1, v_2,..v_k$) of objects. Furthermore, the RNN take these feature representations of objects as input to construct the parsing tree, where $v_i$ is mapped into a semantic representation $x_i$.}
\label{fig:inference}
\vspace{-2mm}
\end{figure*}

Most of the existing scene labeling / parsing models are studied in the context of supervised learning, and they rely on expensive annotations. To overcome this issue, one can develop alternative methods that train the models from weakly annotated training data, e.g., image-level tags and contexts \cite{DBLP:Weakly2}\cite{DBLP:Weakly-MultiInstance}\cite{DBLP:WeaklySegmentation}. Among these methods, one inspiring us is \cite{DBLP:WeaklySegmentation}, which adopts an EM learning algorithm for training the model with image-level semantic labels. This algorithm alternates between predicting the latent pixel labels subject to the weak annotation constraints and optimizing the neural network parameters.

\section{CNN-RNN Architecture}
\label{CNN-RNN Architecture}

Structured scene parsing aims to infer the following three forms of outputs from an image: i) the location of semantic entities, ii) interaction relations and iii) the hierarchical configuration among the semantic entities. To this end, we propose a novel deep architecture by integrating the convolutional neural network (CNN) and recursive neural network (RNN). In our CNN-RNN architecture, the CNN model is introduced to perform semantic segmentation by assigning an entity label (i.e. object category) to each pixel, and the RNN model is introduced to discover hierarchical structure and interaction relations among entities.

Fig.~\ref{fig:inference} illustrates the the proposed CNN-RNN architecture for structured scene parsing. First, the input image is directly fed into our revised VGG-16 network~\cite{vggnet} to produce a score map for each entity category. Based on the softmax normalization of the score maps, each pixel is labeled with an entity category. We further group the adjacent pixels with the same label into an object, and obtain the feature representations of objects. By feeding these feature representations of entities to the RNN, a greedy aggregation procedure is implemented for  constructing the parsing tree. In each recursive iteration, two input objects (denoted by the child nodes) are merged into a higher-level object (denoted the parent node). The finally generated root note represents the whole scene. Different from the RNN architecture in~\cite{DBLP:Recursive_Socher}\cite{DBLP:RecursiveContext2}, our model predicts the relation between these two nodes when they are combined into a higher-level node.

In the following, we discuss the CNN and RNN models in details.

\subsection{CNN Model}
\label{sub:cnn_model}

The CNN model is designed to accomplish two tasks: semantic labeling and generating feature representations for entities. For semantic labeling, we adopt the fully convolutional network with parameters $W_C$ to yield $K+1$ score maps $\{{s}^0, ..., {s}^k, {s}^K\}$, corresponding to one extra background category and $K$ object categories. The score $s_j^k$ is further normalized using softmax to obtain the corresponding classification score:
\begin{equation}\label{eq_softmax}
\sigma(s_j^t) = \frac{\exp(s_j^t)} {\sum_{k=1}^K \exp(s_j^k)}
\end{equation}
where $\sigma(s_j^t)$ denotes the probability of $j$-th pixel belonging to $t$-th object category with $\sum_{t=1}^K \sigma(s_j^t)=1$. $C=\{c_j\}_{j=1}^M$ denotes the labels of pixels in the image $I$, where $c_j\in \{1,...,K\}$ and $M$ is the number of pixels of image $I$. With $\sigma(s_j^t)$, the label of the $j$-th pixel can be predicted by:
\begin{equation}\label{eq_prediction}
c_j = \arg\max_t \: \sigma(s_j^t)
\end{equation}

For generating feature representation for each entity category, we group the adjacent pixels with the same label into a semantic entity category.

Considering that the pixel numbers vary with the semantic entity categories, in order to obtain feature representation with fixed length for any entity category, we use \textit{Log-Sum-Exp}(LSE)~\cite{boyd2004convex}, a convex approximation of the \textit{max} function, to fuse the features of pixels
\begin{equation}\label{eq_feature}
v_k = \frac{1}{\pi}\log \left[\frac{1}{Q_k}\sum_{c_j = k} \exp (\pi \bar{v}_j) \right]
\end{equation}
where $v_k$ denotes the feature representation of the $k$-th entity category, $\bar{v}_j$ denotes the feature representation of the $j$-th pixel by concatenating all feature maps at the layer before softmax at position $j$ into a vector, $Q_k$ is the total number of pixels of the $k$-th object category, and $\pi$ is a hyper-parameter to control smootheness. With higher value of $\pi$, the function tend to preserve the max value for each dimension in the feature, while with lower value the function behaves like a averaging function.

\subsection{RNN Model}
\label{sub:rnn_model}

\begin{figure}[t]
\centering
\includegraphics[width= 2.4in]{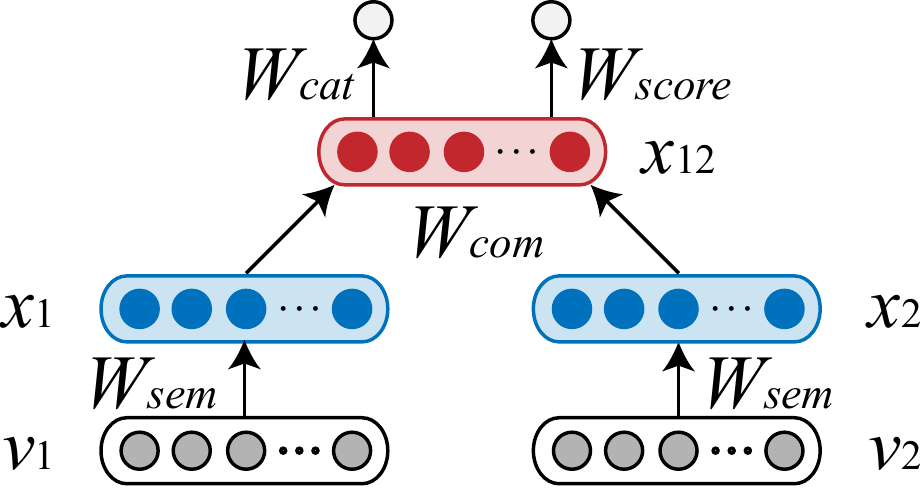}

 \caption{An illustrate of recursive neural network in our CNN-RNN architecture. This network calculates the score for merging decision and predicts the relation category of two merged regions.}
\label{fig:RNN}
\end{figure}


With the feature representations of object categories produced by CNN, the RNN model is designed to generate the image parsing tree for predicting object interaction relations and hierarchical scene structure. The RNN model consists of four sub-networks: (\textit{semantic mapper}, \textit{combiner}, \textit{categorizer} and \textit{scorer}). Therefore, the parameters of the RNN also includes four parts, denoted as $W_R=\{W_{sem},W_{com},W_{cat},W_{score}\}$.

Following~\cite{DBLP:Recursive_Socher} and~\cite{DBLP:RecursiveContext2}, object feature $v_k$ produced by CNN is first mapped onto a semantic space by the \textbf{Semantic mapper}, which is a one-layer fully-connected network.
\begin{equation}\label{eq_sem}
x_k = F_{sem}(v_k;W_{sem})
\end{equation}
where $x_k$ is the mapped feature, $F_{sem}$ is the network transformation and $W_{sem}$ is the network parameter.

The features of two child nodes are fed to the \textbf{Combiner} and generate their parent node feature.
\begin{equation}\label{eq_com}
x_{kl} = F_{com}([x_k,x_l];W_{com})
\end{equation}
where $x_k$ and $x_l$ indicate the two child features and $x_{kl}$ denotes their parent feature in the parsing tree. $F_{com}$ is the network transformation and $W_{com}$ is the corresponding parameter. Parent node feature encode semantic information of the combination of its two child nodes, as well as the structural information of this specific merging operation. The parent node feature has the same dimensionality as the child node feature, allowing the procedure can be applied recursively and eventually the root feature can be used to represent the whole image.

When two nodes are merged into a parent node, the \textbf{Categorizer} sub-network determines the relation of these two nodes. Categorizer is a softmax classifier that takes parent node feature $x_{kl}$ as input, and predict the relation label $y_{kl}$,
\begin{equation}\label{eq_rel_category}
y_{kl} = softmax(F_{cat}(x_{kl};W_{cat}))
\end{equation}
where $y_{kl}$ is the predicted relation probability vector, $F_{cat}$ denotes the network transformation and $W_{cat}$ denotes the network parameter.

The \textbf{Scorer} sub-network measures the confidence of a merging operation between two nodes. It takes the parent node feature $x_{kl}$ as input and outputs a real value $h_{kl}$.
\vspace{-1mm}
\begin{equation}\label{eq_node_score}
h_{kl} = F_{score}(x_{kl};W_{score})
\end{equation}
where $F_{score}$ denotes the network transformation and $W_{score}$ denotes the network parameter. The merging score $q_{kl}$ of node $\{kl\}$ is computed as $q_{kl} = \frac{1}{1+exp(h_{kl})}$.

Merging score is used to optimize the structure discovery in training, as described in Sect. \ref{sub:rnn_loss}.

Similar to \cite{DBLP:Recursive_Socher}, we use the RNN model to construct the parsing tree with a greedy algorithm. The procedure begins with a initial set of leaf nodes. In each iteration, the algorithm enumerates all possible merging pairs and computes merging scores for each. The algorithm chooses the pair with highest score to merge, replacing the pair of nodes with their parent node. The algorithm iterates until there is only one root node left.

\section{Weakly-supervised Model Training}

Compared with some other weak annotations such as labels and attributes, sentences usually provide richer semantics and structured contexts (\eg, object interactions and relations). More importantly, describing images by sentences finely accords with the process of human perception, and it thus contributes to meaningful representation learning.

In the initial stage of model training, we first convert each sentence into a normalized tree by using common techniques, as discussed above. Formally, a semantic tree $T$ includes entity labels (\ie, nouns) and the relations (\ie, verb phrases).

Since the scene configurations are unavailable for the training images, we need to estimate them to training our CNN and RNN. Thus, we train the model with a EM type algorithm. This algorithm alternates between predicting the latent scene configurations (via transferring knowledge from the semantic trees), and optimizing the neural network parameters.

Our model performs two tasks: semantic labeling and scene structure discovery. Thus we define the loss function as the sum of two terms: semantic label loss $\mathcal{J}_{C}$ produced by CNN, and scene structure loss $\mathcal{J}_{R}$ produced by RNN. With a training set containing $Z$ image-tree pairs $\{(I_1,T_1),...,(I_Z,T_Z)\}$. The overall loss function is as follows,
\begin{equation}\label{eq_overallloss}
\mathcal{J}(W)  =  \frac{1}{Z} \sum_{i=1}^Z ( \mathcal{J}_{C}(W_C;I_i,T_i) + \mathcal{J}_{R}(W;V_i,T_i) )
\end{equation}
where $I_i$ is the $i$-th image and $T_i$ is the tree sructure produced from the descriptive sentence. $V_i$ is the set of semantic entity features produced by CNN from the $i$-th image. $V$ takes the form $V=\{v_k|k\in \psi(T)\}$, where $\psi(T)$ is set of object categories mentioned in $T$. $W$ is all model parameters, $W_C$ is model parameters of the CNN.


\begin{figure}[t]
\centering
\includegraphics[width= 3.4 in]{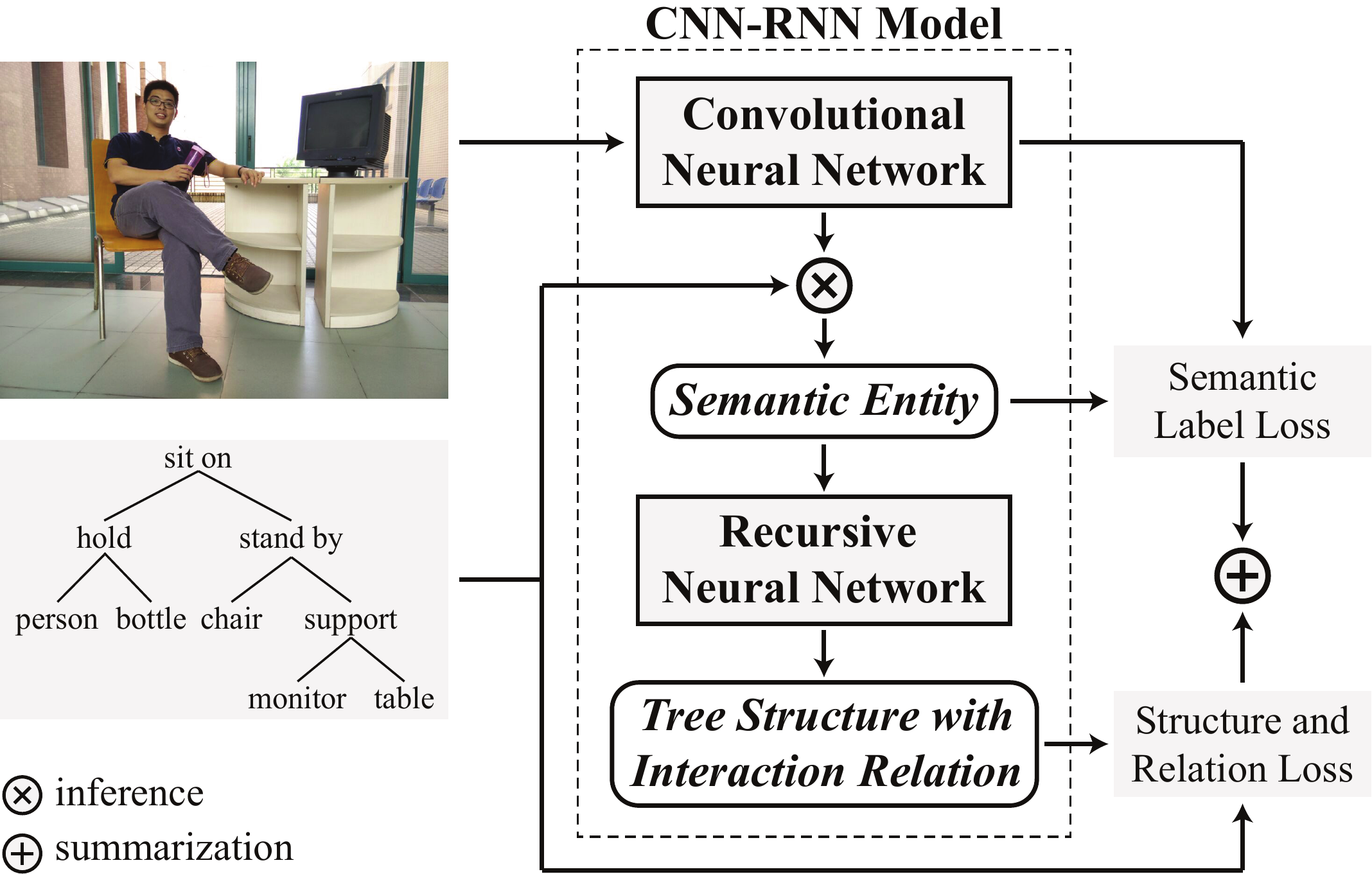}
\caption{An illustration of the training process with our CNN-RNN architecture. The learning objective consists of two proportions: the semantic object labeling via the CNN, and the structure prediction via the RNN.}
\label{fig:learning}
\end{figure}

\subsection{Semantic Label Loss} \label{sub:cnn_loss}
Given intermediate label map $C$, the semantic label task performed by CNN can be optimized as a pixel-wise classfication problem. We first perform an inference step to obtain an estimated ground truth label map $\widehat{C}$, which is used as supervision (see Sect.~\ref{sub:em_learning} for more details). Let $\widehat{c}_j \in \widehat{C}$ denote the estimated category label of pixel $j$, the loss function of semantic labeling for image $I$ is defined as,

\begin{equation}\label{eq_cnn_loss}
\begin{split}
&\mathcal{J}_{C}(W_C;I,T) = -\frac{1}{M}( \sum_{j=1}^M \sum_{k=1}^K \textbf{1}(\widehat{c}_j=k)\log \sigma(s_j^k)\\
& +(1-\textbf{1}(\widehat{c}_j=k))\log(1-\sigma(s_j^k)) ) + \|W_C\|^2
\end{split}
\end{equation}
where $M$ denotes the total number of pixels in the image $I$. As defined in Eq.(\ref{eq_softmax})function $\sigma(s_j^k)$ outputs the probability of $j$-th pixel for the $k$-th entity category predicted by the CNN. Note that $\{s^0,...,s^K\}$ represent the score maps of image $I$ produced by the fully convolutional network with parameters $W_C$.

\subsection{Scene Structure Loss} \label{sub:rnn_loss}
The scene structure discovery task is performed by the RNN, and can be further divided into two sub-tasks: tree structure construction and relation categorization. Thus we define the RNN loss to be the sum of loss from the two tasks,
\begin{equation}\label{eq_rnn_loss}
	\mathcal{J}_{R}(W;V_i,T_i) = \mathcal{J}_{struc}(W;V_i,T_i) + \mathcal{J}_{rel}(W;V_i,T_i)
\end{equation}

\textbf{Tree Structure Construction.}
The goal of tree structrue construction is to learn a transformation $I\rightarrow \mathcal{P}_I$ according to the tree structure $T$. We define an image parsing tree as valid if the sequence of two regions merges is consistent with the merging order in the text parsing tree. From a valid parsing tree, we extract a sequence of ``correct'' merging operations as $\mathcal{A}(V,T)=\{a_1,...,a_{P_T}\}$. $P_T$ is the total number of merging operation in the text parsing tree $T$. This implies a contraint that the nubmer of merging operation in a tree structure always equals nubmer of merging operation in the corresponding text parse tree.

We define a loss based on the merging score $q$ produced by scorer sub-network as described in Sect.~\ref{sub:rnn_model}. For convenience, we denote merging score of operation $a$ given $V$ and $T$ as $q(a)$. Intuitively, we encourage the correct merging operation $a$ to have a larger merging score than that of incorrect merging operation $\widehat{a}$. Thus we have $q(a)\geq q(\widehat{a}) + \triangle$, where $\triangle$ is a constant margin. We define the loss function for scene structrue discovery as,
\vspace{-1mm}
\begin{equation}\label{eq_structure_loss}
\begin{split}
\mathcal{J}_{struc}(W;V,T) & =  \frac{1}{P_T} \sum_{p=1}^{P_T} [ \:\: \max_{\widehat{a}_p\notin\mathcal{A}(V,T)}
q(\widehat{a}_p)\\
& - q(a_p) + \triangle \:] + \frac{\lambda}{2}||W||^2
\end{split}
\end{equation}
where $\lambda$ is the weight of regularization term. Intuitively, this loss objective function maximizes the score of correct merging operation and minimizes incorrect merging operations. To improve efficiency, we do not minimize all incorrect merging operations, but only the one with highest score.

\textbf{Relation Categorization.}
The relation categorization task can be optimized as a softmax classification problem. We define the object function of relation categorization for image $I$ as,
\begin{equation}\label{eq_relation_loss}
\begin{split}
& \mathcal{J}_{rel}(W;V,T) = -\frac{1}{|U_T|}(  \sum_{\{kl\}} \sum_{s=1}^S \textbf{1}(r_{kl}=s)\log G_s(\theta_{kl}(V,W))  \\
& +(1-\textbf{1}(r_{kl}=s))\log(1-G_s(\theta_{kl}(V,W))  ) + \|W\|^2
\end{split}
\end{equation}
$|U_T|$ denotes the number of relation appearing in the tree structre $T$. $\{kl\}$ denotes a node merged from node $k$ and $l$. $S$ is the total number of relation categories. $r_{kl}$ denotes the ground truth relations provided by tree structure $T$ between two semantic entities. $G_s(\theta_{kl}(V,W))$ is the categorizer sub-network in the RNN(see Sect.~\ref{sub:rnn_model}), which outputs the probability that node $\{kl\}$ belongs to relation category $s$.

\subsection{Learning Algorithm} \label{sub:em_learning}

The Expectation-Maximization method is adopted to optimize the loss in Eq.\eqref{eq_overallloss}. In the E-step, guided by the sentence description, we update the intermediate label maps $C$ and the latent structured configurations together with the CNN and RNN losses. In the M-step, the parameters are updated using the back-propagation algorithm. In summary, our learning algorithm can be conducted by iteratively performing the following tree steps:


\textbf{(i) Updating intermediate label maps $\hat{C}$ and the CNN loss.} Given image $I$ and its semantic tree $T$, we compute the classification probability of each pixel according to Eq.\eqref{eq_softmax}. Inspired by the work of cardinality potentials~\cite{DBLP:conf/uai/TarlowSZAF12}\cite{DBLP:conf/icml/LiZ14}, the score of pixel $j$ belonging to the label $k$ is calculated by $f_j(k) = \sigma(s_j^k)+\delta_k$, where $\sigma(s_j^k)$ is defined in Eq.\eqref{eq_softmax}. $\delta_k$ is entity-dependent biases, which is set adaptively according to the prescribed proportion areas of background or foreground entity classes in the image~\cite{DBLP:WeaklySegmentation}, regarding the set of entities in $T$. The final classification result of pixel $j$ is computed by $\widehat{c}_j=\arg\max_k f_j(k)$. Finally, the CNN loss is computed according to Eq.\eqref{eq_cnn_loss}.

\textbf{(ii) Updating latent scene structures and the RNN loss.} Given the label of each pixel, we group the pixels into semantic objects and obtain the object feature representations with the method described in Sect.~\ref{sub:cnn_model}. Then we use the RNN model to infer the interaction relations and hierarchical configuration of objects, and compute the RNN loss according to Eq.\eqref{eq_structure_loss} and Eq.\eqref{eq_relation_loss}.

\textbf{(iii) Updating the CNN and RNN parameters.}  Given the intermediate label maps and latent scene structure, we can compute the gradient of the overall loss in Eq.\eqref{eq_overallloss} w.r.t. the CNN and RNN parameters. With the BP algorithm, the gradients from the semantic label loss propagate backward through all layers of CNN. The gradients from the scene structure loss first propagate recursively through the layers of RNN, and then propagate through the object features to the CNN. Thus, all the parameters of our CNN-RNN model can be learned in an end-to-end manner.

\section{Experiment}
\label{sec:experiments}
We first introduce the implementation details and then evaluate the performance of our proposed method for semantic labeling and structured scene parsing.


\textbf{Datasets.}
We conduct our experiments on PASCAL VOC 2012 segmentation benchmark \cite{pascal_voc}, which contains 4,369 images from three subsets: training (1,464 images), validation (1,449 images) and test(1,456 images). PASCAL VOC 2012 dataset has 20 foreground categories and 1 background category. To suit our task, we randomly divide images in the training and validation sets into 5 groups, and asked 5 annotators to provide one description for each image in each group respectively. Since the groundtruth labeling is unavailable for test images, we did not annotate the test set. In the semi-supervised experiments, the training set is further divided into two subsets, where one is the strongly-annotated subset and the other is the PASCAL VOC 2012 training set with sentence description. Considering the Semantic Boundaries Dataset (SBD)~\cite{sbd_dataset} provides pixel-wise labels for images from PASCAL VOC 2011, we use part of the SBD to constitute the strongly-annotated subset, which includes at most 1,464 of the 10,582 training images in our experiments.

\textbf{Annotation.}
Direct annotation of the structured parsing trees for images is time-consuming, since it requires carefully designed tools and user interface. To save annotation cost, we use the natural language descriptions instead of trees. The sentence description of an image naturally provides a tree structure to indicate the major objects along with their interaction relations~\cite{DBLP:journals/ml/Elman91}. Here we use the Stanford Parser~\cite{DBLP:conf/acl/SocherBMN13} to parse sentences and produce constituency trees, which are two-way trees with each word in a sentence as a leaf node and can serve as suitable alternative of structured image tree annotation.

\begin{figure}[t]
\centering
\includegraphics[width=3.5in]{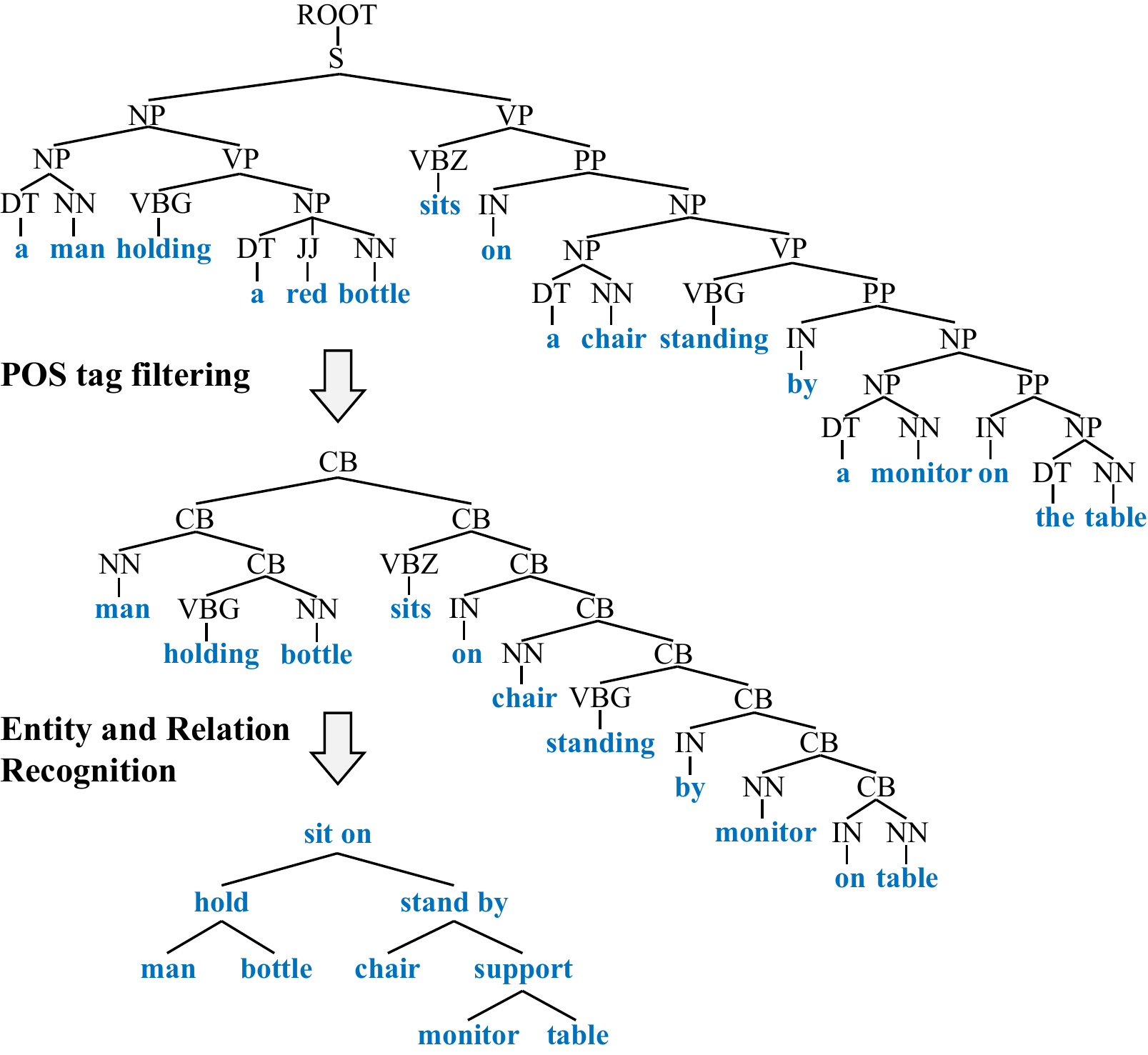}
\caption{An illustration of the tree conversion process. The top tree is the constituency tree generated by language parser. The middle tree is the constituency tree after POS tag filtering. The bottom tree is the converted relation tree.}
\label{fig:language}
\end{figure}

\textbf{Preprocessing.}
Constituency trees from the Stanford Parser~\cite{DBLP:conf/acl/SocherBMN13} still contains irrelevant words that do not describe object category or interaction relations(\eg, adjectives). Therefore, we need to convert constituency trees into semantic trees, which only contains semantic entities and scene structure (as illustrated in Fig.~\ref{fig:language}).

The conversion process generally involves three steps. Given a constituency tree (top tree in Fig.~\ref{fig:language}), we first filter the leaf nodes by their part-of-speech, preserving only nouns as object candidates, and verbs and prepositions as relation candidates. Second, nouns are combined and converted to object categories. Annotators sometimes use different nouns for the same category (\eg ``cat'' and ``kitten''). Thus we use the lexical relation data in WordNet~\cite{wordnet} to unify the synonyms belonging to same defined category, and convert them to the corresponding object category. Annotators may mention entities that are not in any defined object categories (\eg ``grass'' in ``a sheep stands on the grass''), which will be also removed from the trees.

Third, relations should also be recognized and refined. Denote by $R$ a set of defined relations, and $T$ the triplets in the form of $(entity1, verb/prep, entity2)$. We construct a mapping $T \rightarrow R$ to recognize relation. $R$ also contains two special relation categories: ``other'' and ``background''. The ``other'' serves as a placeholder for undefined relations. The ``background'' deals with the special cases where only one entity is recognized in a tree. In this case we merge the entity with an additional ``background'' entity, and assign ``background'' relation to their parent node.

\subsection{Semantic Labeling}
\label{sub:semantic_Labeling}

\begin{figure}[tbp]
\centering
\includegraphics[width=3.4in]{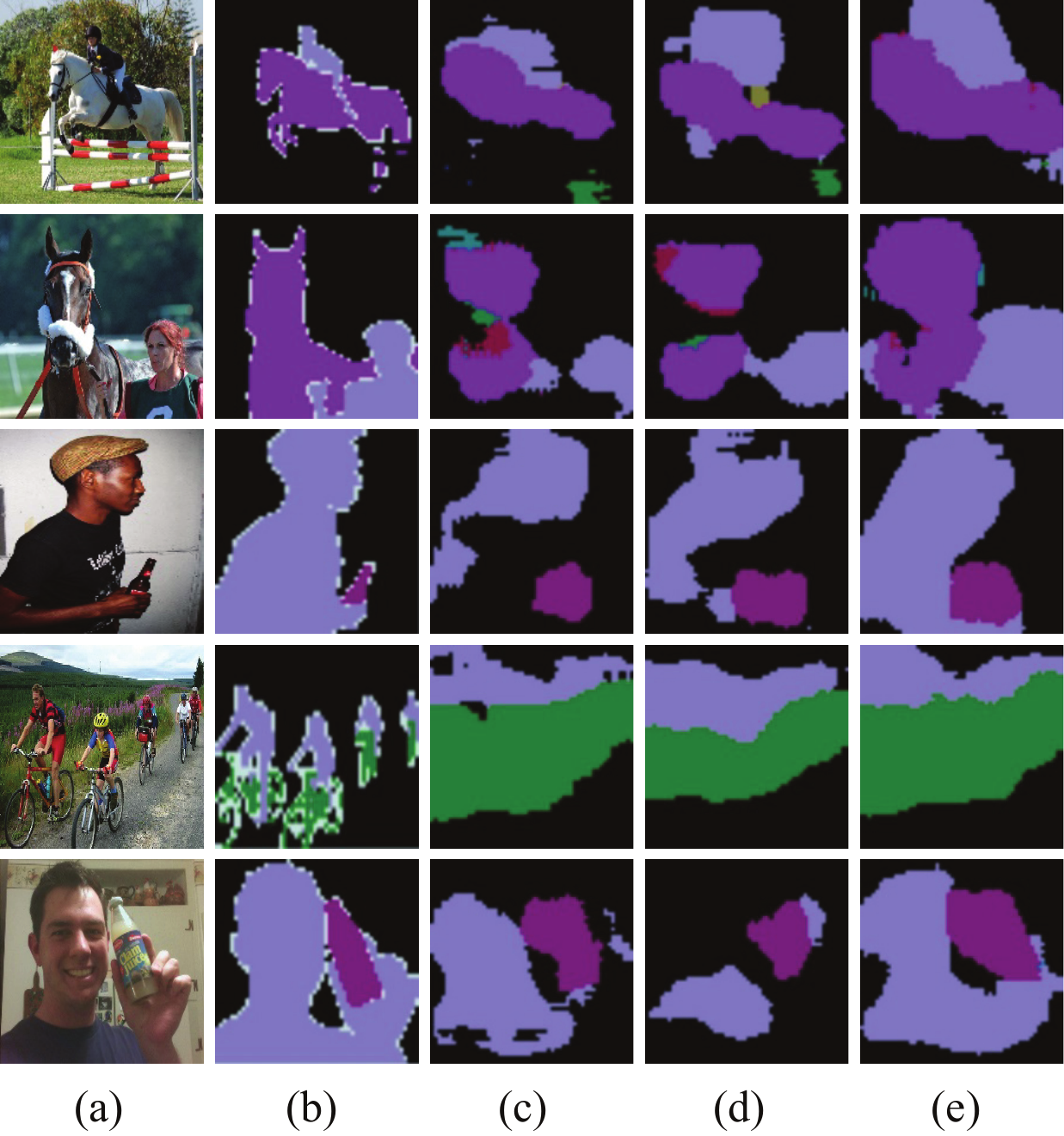}
\caption{Visualized semantic labeling results. (a) The input images; (b) The groundtruth lebeling results; (c) Our proposed method (weakly-supervised); (d) Deeplab(weakly-supervised)~\cite{DBLP:WeaklySegmentation}; (e)MIL-ILP(weakly-supervised)~\cite{pinheiro2015image}}
\label{fig:labeling_result}
\end{figure}

In this section, we report the results for the conventional semantic labeling task which assigns semantic label to each pixel. We consider two experimental settings, \ie weakly-supervised learning and semi-supervised learning, and adopt the pixel-wise intersection-over-union(IoU) used in PASCAL VOC segmentation challenge~\cite{pascal_voc} as the performance indicator. Note that our description annotation does not cover the exact same object classes in each image as in the pixel-wise annotation, making only partial class labels are used for training. For fair comparison, we modified the training and validation images by assigning background category to the object categories not mentioned in description sentences. Due to the labels of the test set is not available, we cannot modify the test set  and thus only report the results on the modified validation set. Visualized labeling results are shown in Fig.~\ref{fig:labeling_result}.

\textbf{Weakly-supervised Learning.}
Table \ref{tbl:result_semantic_segmentation} shows the results under the setting of weakly-supervised learning. We compare our method with MIL-ILP~\cite{pinheiro2015image}, MIL-FCN~\cite{DBLP:FCnetwork}, and DeepLab~\cite{DBLP:WeaklySegmentation}, a state-of-the-art weakly-supervised method using image labels as supervision. We perform experiments with the publicly available code of DeepLab, and our own implementation of MIL-ILP and MIL-FCN. Our method obtains the IoU of 34.3\%, outperforming DeepLab~\cite{DBLP:WeaklySegmentation} by 4\%. If we fix the parameters of the RNN with random initialization, a 2.6\% drop of IoU is observed, indicating that the RNN does help in learning the CNN. 

\begin{table}[!h]\small
\begin{center}
\begin{tabular}{|c|c|}
\hline
Method & IoU \\
\hline
MIL-ILP~\cite{pinheiro2015image} & 29.4\% \\
\hline
MIL-FCN~\cite{DBLP:FCnetwork} & 28.3\% \\
\hline
DeepLab(weakly)~\cite{DBLP:WeaklySegmentation} & 30.3\% \\
\hline
Ours(fixed-RNN) & 31.7\% \\
\hline
Ours & \textbf{34.3}\% \\
\hline

\end{tabular}
\end{center}
\caption{PASCAL 2012 val result of weakly supervised methods}
\label{tbl:result_semantic_segmentation}
\end{table}

\textbf{Semi-supervised Learning.}
In this setting, we have access to both pixel-level (strongly) annotated data and image-level (weakly) annotated data, and our method can take advantage of both types of supervision information in the training procedure. We consider two semi-supervised strategies: waterfall and fusion. For the waterfall strategy, we first perform 8,000 iterations of strongly-supervised pre-training on the CNN, followed by 16,000 iterations of weakly-supervised training on the CNN and RNN. For the fusion strategy, we use a weighted sum of strongly-supervised and weakly-supervised loss functions to train the CNN and RNN, where we use $280$ strong samples together with weak training samples, and the loss weight is set as 1:1 (strong:weak). Table \ref{tbl:semi_supervision} shows the result on the PASCAL VOC 2012 validation set. We observe that all methods benefit significantly from semi-supervised learning. The improvement of IoU compared to weakly supervised learning is 8.9\% with 280 strongly annotated samples (strong:weak = 1:5), and is 16.6\% with 1464 strongly annotated samples (strong:weak = 1:1). Our method outperforms DeepLab~\cite{DBLP:WeaklySegmentation} by 0.7\% with 280 strong samples and fusion strategy.

Given the same number of strongly annotated data, the fusion strategy outperforms the waterfall strategy by 10.2\% in terms of IoU. We observe that the accuracy of pre-training step in waterfall strategy is very high (over 95\%) on the training set. This indicates that the separated pre-training with small amount of data causes the model overfitted, making pre-training contribute little to performance improvement. Nevertheless, the fusion strategy trains the model with a combined loss for better tradeoff of the two types of supervision information, and thus can exploit the strongly annotated data without suffering from overfitting.

\begin{table}[!h]\small
\begin{center}
\begin{tabular}{|c|c|c|c|c|}

\hline
Method & \# strong & \# weak & IoU \\
\hline
MIL-ILP(fusion)~\cite{pinheiro2015image} & 280 & 1464 & 39.3\% \\
\hline
MIL-FCN(fusion)~\cite{DBLP:FCnetwork} & 280 & 1464 & 38.4\%\\
\hline
DeepLab(fusion)~\cite{DBLP:WeaklySegmentation} & 280 & 1464 & 42.5\% \\
\hline
Ours(fusion) & 280 & 1464 & 43.2\% \\
\hline
Ours(fusion) & 1464 & 1464 & 50.9\% \\
\hline
Ours(waterfall) & 280 & 1464 & 33.0\% \\
\hline

\end{tabular}
\end{center}
\caption{PASCAL 2012 val result with semi-supervised learning}
\label{tbl:semi_supervision}
\vspace{-3mm}
\end{table}

\subsection{Structured Scene Parsing}
\label{sub:structure_semantic_parsing}

In this section, we evaluate the structured scene parsing performance of the proposed method, which is measured with two metrics: relation accuracy and structure accuracy. Relation accuracy is computed recursively. Denote by $T$ a tree and $P = \{T, T_1, T_2, \ldots, T_m\}$ the set of enumerated sub-tress (including $T$) of $T$. A leaf $T_i$ is considered to be correct if it is of the same object category as the one in the ground truth tree. A non-leaf $T_i$ (with two subtrees $T_l$ and $T_r$) is considered to be correct if and only if $T_l$ and $T_r$ are both correct and the relation label $r_T$ is correct. Then, the relation accuracy is calculated as $\frac{(\# of correct subtrees)}{m+1}$, and the structure accuracy is a simplification of the relation accuracy by ignoring the relation labels in the evaluation of the correctness of $T$.

Note that not all images in the PASCAL VOC 2012 validation set can be used for structure and relation accuracy, \eg the images containing only one object, and these images should not be counted in the experiments.







To get detailed understanding of our method, we study the effect of two factors, \ie joint CNN/RNN learning and end-to-end learning, and conduct experiments with the following configurations: i) Fixed the other parameters of the CNN except for the top two layers, we update all parameters of the RNN; ii) Fixed all parameters of RNN with randomly initialized values, we update all parameters of the CNN; iii) We separate the learning of CNN and RNN, \ie we first update the CNN for 16000 iterations with the fixed RNN, and then update RNN for 16000 iterations with the fixed CNN; iv) We update both CNN and RNN in the whole process with an end-to-end and joint learning manner.

\begin{table}[!ht]\small
\begin{center}
\begin{tabular}{|c|c|c|c|c|}

\hline
CNN & RNN & struct. acc & rel. acc \\
\hline
partial fixed & updated & 57.0\% & 49.0\% \\
\hline
updated & fixed(rand init) & 40.8\% & 31.4\% \\
\hline
learnt \& fixed & updated & 60.8\% & 54.2\% \\
\hline
updated & updated & 64.2\% & 62.8\% \\
\hline

\end{tabular}
\end{center}
\caption{PASCAL 2012 result with different learning strategies}
\label{tbl:result_end_to_end_learning}
\vspace{-2mm}
\end{table}

Table \ref{tbl:result_end_to_end_learning} shows the results on the PASCAL VOC 2012 validation set. Our method with end-to-end and joint learning performs best among all training settings. The training setting with fixed RNN performs much worse than one with fixed CNN, indicating that the RNN plays a more important role for structure and relation prediction. This is reasonable since structure and relation is finally obtained by RNN. Learning CNN and RNN separately performs better than learning with either fixed, but is still worse than end-to-end and joint learning.

\section{Conclusion}

We have introduced a structured scene parsing method based on a deep CNN-RNN architecture, and a cost-effective mode training method by transferring knowledge from image-level descriptive sentences. We have demonstrated the effectiveness of our framework by i) generating hierarchical and relation-aware configurations from the scene images and ii) achieving more favorable scene labeling results compared to other state-of-the-art weakly-supervised methods.

There are several directions in which we intend to extend this work, such as improving our system by adding a component for object attribute parsing. Deeply combining with some language processing techniques also would be a possible way.

\section*{Acknowledgment}

This work was supported in part by Special Program for Applied Research on Super Computation of the NSFC-Guangdong Joint Fund (the second phase), in part by Guangdong Natural Science Foundation under Grant S2013050014548, in part by Program of Guangzhou Zhujiang Star of Science and Technology under Grant 2013J2200067, and in part by the Fundamental Research Funds for the Central Universities.

{\small
\bibliographystyle{ieee}
\bibliography{egbib}
}

\end{document}